\documentclass{article} 
\usepackage{iclr2025_conference,times}

\usepackage{amsmath,amsfonts,bm}

\def\eqref#1{equation~\ref{#1}}

\def\1{\bm{1}}

\DeclareMathAlphabet{\mathsfit}{\encodingdefault}{\sfdefault}{m}{sl}
\SetMathAlphabet{\mathsfit}{bold}{\encodingdefault}{\sfdefault}{bx}{n}

\usepackage{hyperref}
\usepackage{url}

\usepackage{graphicx}
\usepackage{xcolor}
\usepackage{colortbl}
\usepackage{amssymb}
\usepackage{multirow}  
\usepackage{amsmath}
\definecolor{mylightgray}{RGB}{220,220,220}
\definecolor{mylightblue}{RGB}{202, 241, 202}
\usepackage{wrapfig}
\usepackage{floatflt} 
\usepackage{graphicx}
\usepackage{placeins}

\title{Assessing the Creativity of LLMs in Proposing Novel Solutions to Mathematical Problems}

\author{Junyi Ye$^{\dagger}$, Jingyi Gu$^{\dagger}$, Xinyun Zhao$^{\dagger}$, Wenpeng Yin$^{\P}$, Guiling Wang$^{\dagger}$ \\
$^{\dagger}$New Jersey Institute of Technology,
$^{\P}$Pennsylvania State University\\
\texttt{\{jy394,jg95,xz43,guiling.wang\}@njit.edu}; \texttt{wenpeng@psu.edu}
}

\newcommand{\dataname}{\textsc{CreativeMath}}

\iclrfinalcopy 
\begin{document}

\maketitle

\begin{abstract}
The mathematical capabilities of AI systems are complex and multifaceted. Most existing research has predominantly focused on the \emph{correctness} of AI-generated solutions to mathematical problems. In this work, we argue that beyond producing correct answers, AI systems should also be capable of, or assist humans in, developing \emph{novel solutions} to mathematical challenges. This study explores the creative potential of Large Language Models (LLMs) in mathematical reasoning, an aspect that has received limited attention in prior research. We introduce a novel framework and benchmark, \dataname, which encompasses problems ranging from middle school curricula to Olympic-level competitions, designed to assess LLMs' ability to propose innovative solutions after some known solutions have been provided. Our experiments demonstrate that, while LLMs perform well on standard mathematical tasks, their capacity for creative problem-solving varies considerably. Notably, the Gemini-1.5-Pro model outperformed other LLMs in generating novel solutions. This research opens a new frontier in evaluating AI creativity, shedding light on both the strengths and limitations of LLMs in fostering mathematical innovation, and setting the stage for future developments in AI-assisted mathematical discovery \footnote{Data \& Code: \url{https://github.com/JunyiYe/CreativeMath}}.
\end{abstract}

\section{Introduction}

In recent years, artificial intelligence has made significant strides, particularly in the development of Large Language Models (LLMs) capable of tackling complex problem-solving tasks. Models like GPT-4 and Gemini-1.5-Pro have demonstrated impressive proficiency on rigorous mathematical benchmarks \citep{DBLPAhnVLLZY24} such as GSM8K \citep{cobbe2021training} and MATH \citep{hendrycks2021measuring}, underscoring the evolving role of LLMs from simple text generators to sophisticated tools capable of engaging with high-level mathematical challenges. Beyond solving student-oriented math problems, leading mathematicians have begun exploring the use of LLMs to assist in tackling unresolved mathematical challenges \citep{DBLPRomeraParedesBNBKDREWFKF24,DBLPTrinhWLHL24}. Despite these models' success in achieving high accuracy on existing mathematical datasets, their potential for creative problem-solving remains largely underexplored.

Mathematical creativity goes beyond solving problems correctly; it involves generating novel solutions, applying unconventional techniques, and offering deep insights—areas traditionally associated with human ingenuity. Yet, most studies have focused primarily on correctness and efficiency, paying little attention to the innovative approaches LLMs might employ. Furthermore, creativity in mathematical problem-solving is rarely integrated into existing benchmarks, limiting our understanding of LLMs' full potential. The current research landscape lacks a comprehensive framework that evaluates both the accuracy and the creative capacity of LLMs. This gap highlights the need for new methodologies and benchmarks specifically designed to assess and cultivate the creative problem-solving abilities of LLMs in mathematics, which is the focus of this paper.

We created the dataset \dataname, a comprehensive math benchmark that includes problems from middle school to Olympic-level competitions, each accompanied by multiple high-quality solutions ranging from straightforward to highly innovative approaches. Additionally, we designed a multi-stage framework to rigorously evaluate the creativity of LLMs in generating novel math solutions. This evaluation spans closed-source, open-source, and math-specialized LLMs, assessing both the correctness and novelty of their solutions based on different reference prior solutions. 


Our evaluation revealed several interesting key insights: (1) Gemini-1.5-Pro excelled in generating unique solutions, with most correct answers being distinct from the provided references, while smaller and math-specialized models struggled with novelty. (2) Providing more reference solutions generally improved accuracy, with Gemini-1.5-Pro achieving perfect accuracy with four prior solutions. However, increased references made it harder for models to generate unique solutions, indicating a trade-off between leveraging existing knowledge and fostering creativity. (3) As math problem difficulty increased, LLM accuracy declined, but successful solutions were more likely to be innovative, suggesting that tougher problems encourage creativity. (4) Analysis of solution similarity among different LLMs showed that models like Llama-3-70B and Yi-1.5-34B explored diverse approaches, while others like Mixtral-8x22B produced more similar solutions, highlighting the value of using a diverse set of LLMs to enhance originality.

This study lays the groundwork for future advancements in LLM math creativity. The major contributions include: (1) Introducing a new task—evaluating LLMs' mathematical creativity, (2) Creating the \dataname~dataset, (3) Developing a framework for assessing mathematical creativity in LLMs, and (4) Evaluating state-of-the-art LLMs, revealing key insights into their strengths and limitations.


\section{Related Work}

LLMs have demonstrated significant advancements in both mathematical reasoning and creative capabilities, making them increasingly powerful tools in a variety of domains. In the realm of mathematical reasoning, techniques such as prompt engineering, Chain-of-Thought (CoT) prompting, and program-aided language modeling have notably enhanced LLMs' abilities to solve complex problems \citep{brown2020language, wei2022chain, zhou2023solving}. These approaches enable models to break down problems into more manageable steps, thereby improving their accuracy and reasoning depth. Moreover, specialized models like MathVerse \citep{zhang2024mathverse} and Internlm-Math \citep{ying2024internlm}, which are trained on extensive mathematical corpora, have achieved significant improvements in mathematical problem-solving performance \citep{lewkowycz2022solving,ying2024internlm}. Benchmarks such as GSM8K and MATH further provide a structured means to evaluate and compare these advancements, highlighting the continuous progress in this area \citep{cobbe2021training,hendrycksmath2021}.

In terms of creativity, LLMs have shown remarkable prowess across diverse fields. They have excelled in generating high-quality, human-like content, ranging from code generation \citep{ni2023lever,liu2024your} and music composition  \citep{yuan2024chatmusician} to literature  \citep{gomez2023confederacy,liu2024ai} and educational tools  \citep{lan2024teachers,orenstrakh2023detecting}. Creativity in LLMs is often evaluated using frameworks like Margaret Boden's criteria  \citep{franceschelli2023creativity} and the Torrance Tests of Creative Thinking (TTCT)  \citep{torrance1966torrance}, where they have demonstrated high fluency, originality, and flexibility. However, the applicability of these traditional creativity metrics to AI systems is still a topic of debate, as they were originally designed to assess human creativity  \citep{zhao2024assessing}. Techniques such as associative thinking have been employed to enhance the creative output of LLMs further, although challenges remain in ensuring that these models can meaningfully integrate unrelated concepts  \citep{mehrotra2024enhancing}. 
The ethical and legal implications of AI-generated creativity continue to be a significant area of concern, underscoring the need for ongoing research to refine evaluation methods and address societal impacts  \citep{lofstead2023economic}.

\section{\dataname~Curation}

This section details the creation, collection, and processing of our dataset \textbf{CreativeMath}, which comprises high-quality mathematical problems from various competitions and their numerous solutions. The dataset is diverse, encompassing a broad range of mathematical topics and problem types, and covers difficulty levels from middle school to Olympiad level. It includes problems from eight major US competitions: AMC 8, AMC 10, AMC 12, AHSME, AIME, USAJMO, USAMO, and IMO.


\paragraph{Data Collection.}
The dataset was sourced from the Art of Problem Solving (AoPS)\footnote{Art of Problem Solving. ``AoPS Wiki'', \url{https://artofproblemsolving.com/wiki/}.}, a platform offering the most comprehensive collection of problems from various math competitions, along with multiple solutions contributed by participants over the years. As the most popular and sought-after resource for math competitors, AoPS effectively functions as a natural crowdsourcing platform. It uniquely approximates the complete set of viable human solutions for each problem, with later contributors often building on earlier ones.

We meticulously scraped data from eight competitions, ranging from middle school level to Olympic-level, to capture the breadth of mathematical challenges and the depth of solution strategies available. 

\paragraph{Data Cleaning.}
To ensure the integrity and reliability of the dataset, we conducted a rigorous data cleaning procedure. We accurately extracted LaTeX-formatted problems and solutions from HTML, ensuring their correct representation. Irrelevant comments were removed to make each problem and solution clear and self-sufficient. Samples with images, problems without solutions, or incomplete entries were manually removed from the dataset. After this process, the dataset comprises 6,469 mathematical problems and 14,223 solutions. Each problem in the dataset is tagged with detailed metadata, including difficulty level, math category, and problem type. Difficulty levels and problem types were assigned based on official competition data, while the math category were determined using the Llama-3-70B model. 

\begin{figure}[t]
    \centering
    \includegraphics[width=0.9\textwidth]{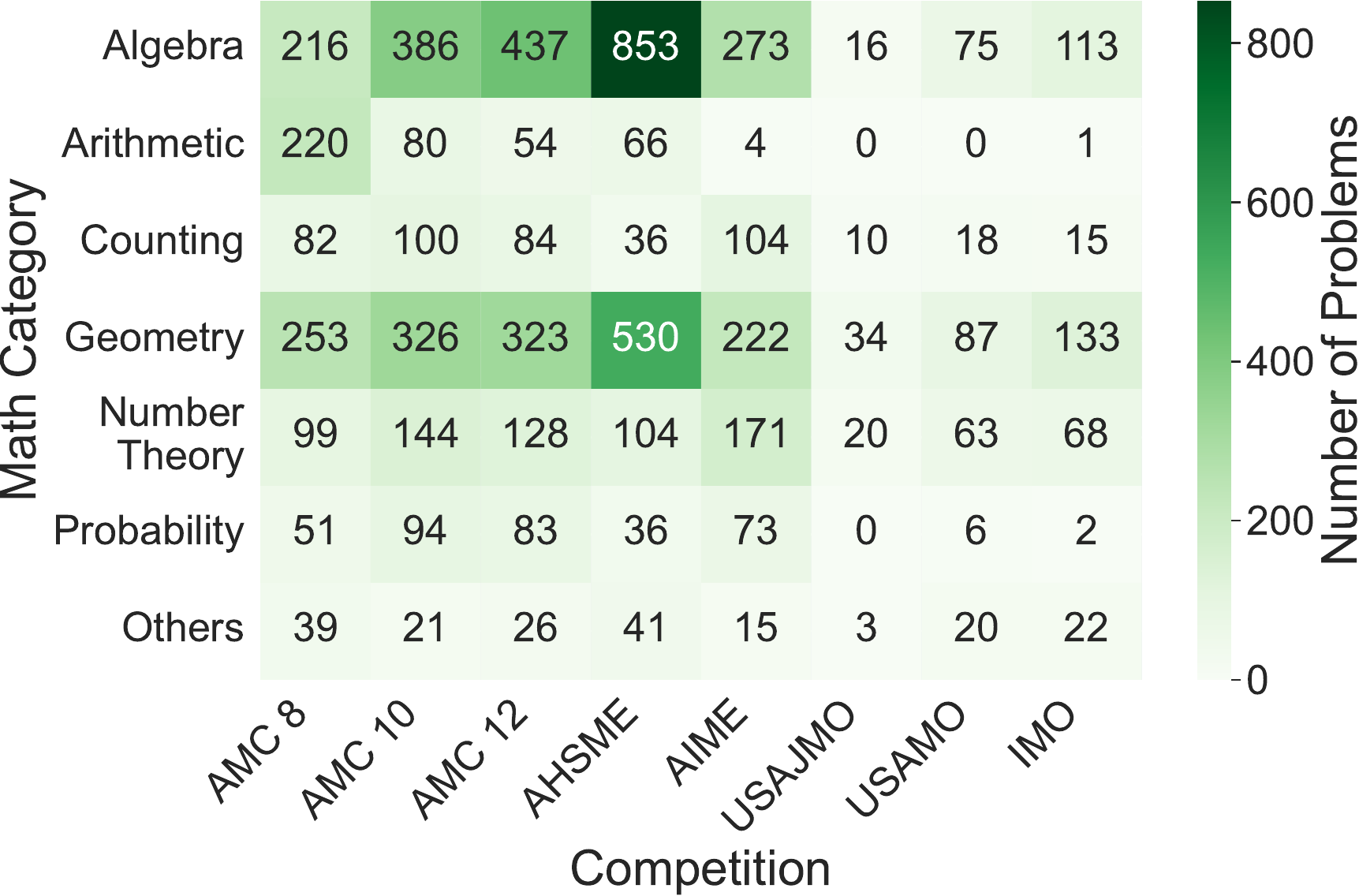}
    \caption{Distribution of problems across different math categories and competitions in the CreativeMath dataset.}
    \label{fig:dataset_distribution}
\end{figure}

\begin{figure}[t]
    \centering
    \includegraphics[width=0.9\textwidth]{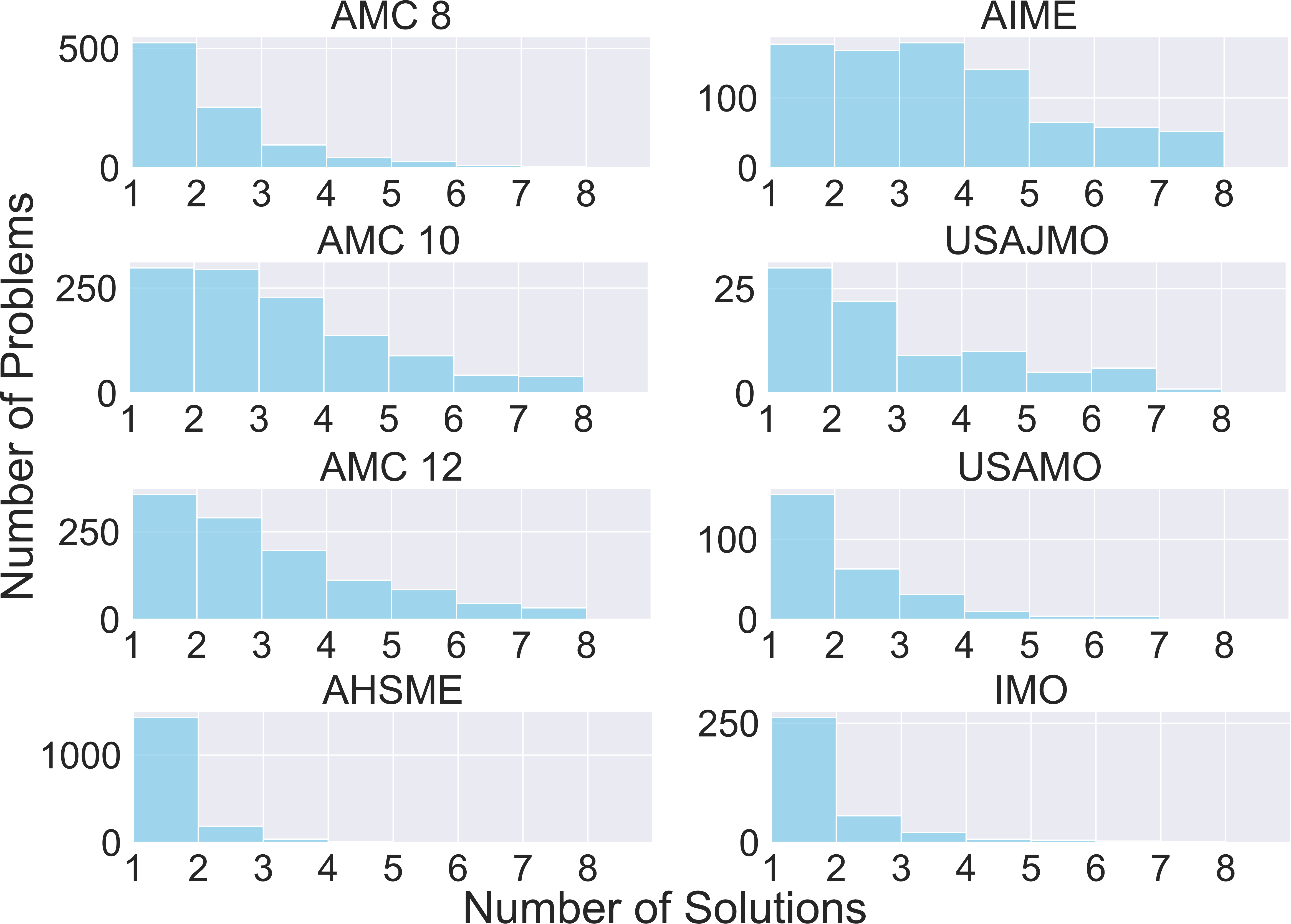}
    \caption{Distribution of the number of solutions per problem across different competitions.}
    \label{fig:number_of_solution_distribution}
\end{figure}

\paragraph{Dataset Analysis.}
As shown in Figure \ref{fig:dataset_distribution}, the problem distribution inside CreativeMath reveals that Algebra and Geometry are the most represented categories across all competitions.
The number of solutions across different competitions, as depicted in Figure \ref{fig:number_of_solution_distribution}, reflects the varying complexity of the problems. Medium-difficulty competitions like AMC 10, AMC 12, and AIME typically have a larger number of solutions, as these problems allow for a variety of approaches. In contrast, simpler competitions like AMC 8 tend to have fewer solutions due to the straightforward nature of the problems, which often have limited methods of solving. Olympic-level competitions such as USAJMO, USAMO, and IMO also see fewer solutions, likely due to the high complexity of the problems, which limits the number of viable solving strategies.



\begin{figure*}[t]
    \centering
    \includegraphics[width=1\textwidth]{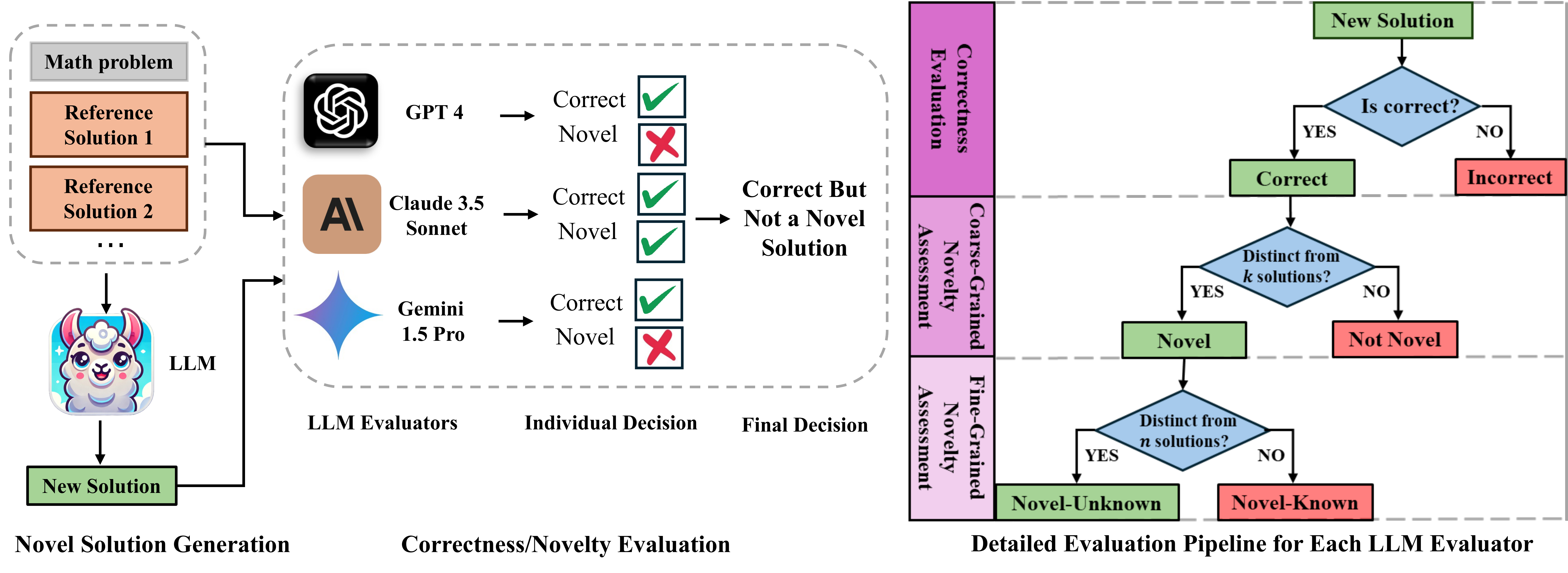}
    \caption{The framework includes solution generation (left) and the evaluation pipeline (middle). The flowchart of the detailed evaluation pipeline is illustrated on the right.}
    \label{fig:framework}
\end{figure*}

\section{Methods} 
Our approach consists of a multi-stage pipeline designed to evaluate the novelty of mathematical solutions generated by an LLM. The methodology is structured into four key stages: \textit{Novel Solution Generation}, \textit{Correctness Evaluation}, \textit{Coarse-Grained Novelty Assessment}, and \textit{Fine-Grained Novelty Assessment}. This comprehensive pipeline illustrated in Figure \ref{fig:framework} ensures that the generated solutions are not only correct but also exhibit a meaningful degree of novelty relative to the reference solutions. The sample prompts and LLMs' responses are provided in the Appendix.


\clearpage

\begin{wrapfigure}[20]{r}{0.5\textwidth}
    \centering
    \vspace{-1.2cm}
    \includegraphics[width=0.5\textwidth]{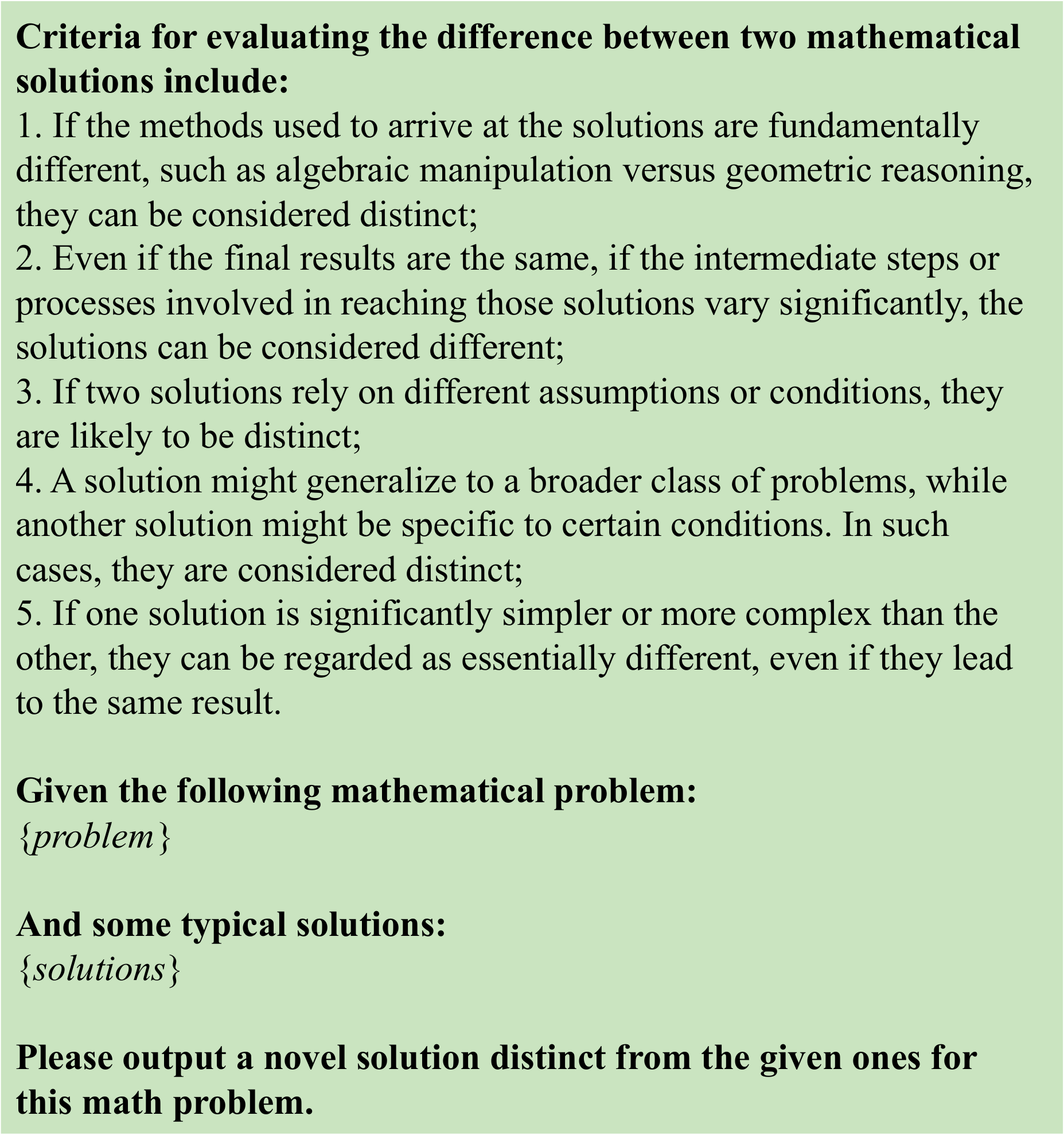}
        \vspace{-0.7cm}
    \caption{The prompt template for generating novel solution.}
    \label{fig:prompt_generation}
\end{wrapfigure}
\subsection{Novel Solution Generation} 
The first stage of the methodology aims to generate novel solutions for the given mathematical problem using LLM. For each problem, a subset of \(k\) reference solutions (where \(k\) ranges from 1 to \(n\), with \(n\) representing the total number of available reference solutions) is sequentially selected based on the order in which competitors uploaded their solutions on the website. 
Earlier solutions are often the most common and intuitive, while later ones may build on previous methods, offer improvements, or introduce entirely novel algorithms.
Consequently, as \(k\) increases, the difficulty in generating new and innovative solutions also increases. 


To ensure clarity and consistency in both prompting and evaluating the novelty of generated solutions, we define a set of criteria agreed upon in consultation with several mathematicians. These criteria guide both the generation and the evaluation process and are used to assess the distinctiveness of the solutions. The criteria are as follows:

\enspace\enspace\textbullet\enspace\textbf{Methodological Differences}: If the methods used to arrive at the solutions are fundamentally different (e.g., algebraic manipulation versus geometric reasoning), the solutions are considered distinct.

\enspace\enspace\textbullet\enspace\textbf{Intermediate Step Variation}: Even if the final results are identical, if the intermediate steps or processes involved in reaching those solutions differ significantly, the solutions are considered novel.

\enspace\enspace\textbullet\enspace\textbf{Assumptions and Conditions}: Solutions that rely on different assumptions, initial conditions, or constraints are treated as distinct.
    
\enspace\enspace\textbullet\enspace\textbf{Generalization}: A solution that generalizes to a broader class of problems is considered novel compared to one that is specific to certain conditions.

\enspace\enspace\textbullet\enspace\textbf{Complexity}: If one solution is notably simpler or more complex than another, they are regarded as different, even if they lead to the same final result.

These criteria, also illustrated in Figure \ref{fig:prompt_generation}, are embedded into the prompt used to guide the LLM in generating novel solutions. The reference solutions provided to the model aim to capture a variety of approaches, and the LLM is instructed to output a new solution that is distinct according to the defined criteria. The prompt emphasizes generating solutions that use different problem-solving methods, distinct intermediate steps, and variations in assumptions or generalizability.

As part of this process, to avoid influencing the judgment of evaluators during the subsequent evaluation stage, transition sentences and justifications explaining why the new solution is distinct from the reference solutions are manually removed. Only the newly generated solution is presented for evaluation.


\subsection{Correctness and Novelty Evaluation}
To rigorously evaluate the correctness and novelty of the generated solutions, we employ three leading LLMs—GPT-4, Claude 3.5 Sonnet, and Gemini 1.5 Pro—as \textbf{LLM Evaluators}, recognized among the strongest models available. These LLM Evaluators collaboratively assess the solutions following the framework illustrated in Figure \ref{fig:framework} (middle). Each LLM Evaluator adheres to the flowchart depicted in Figure \ref{fig:framework} (right) to systematically evaluate the generated solutions across three dimensions:

\begin{wrapfigure}[26]{r}{0.5\textwidth}
    \centering
    \vspace{-1.5cm}
    \includegraphics[width=0.5\textwidth]{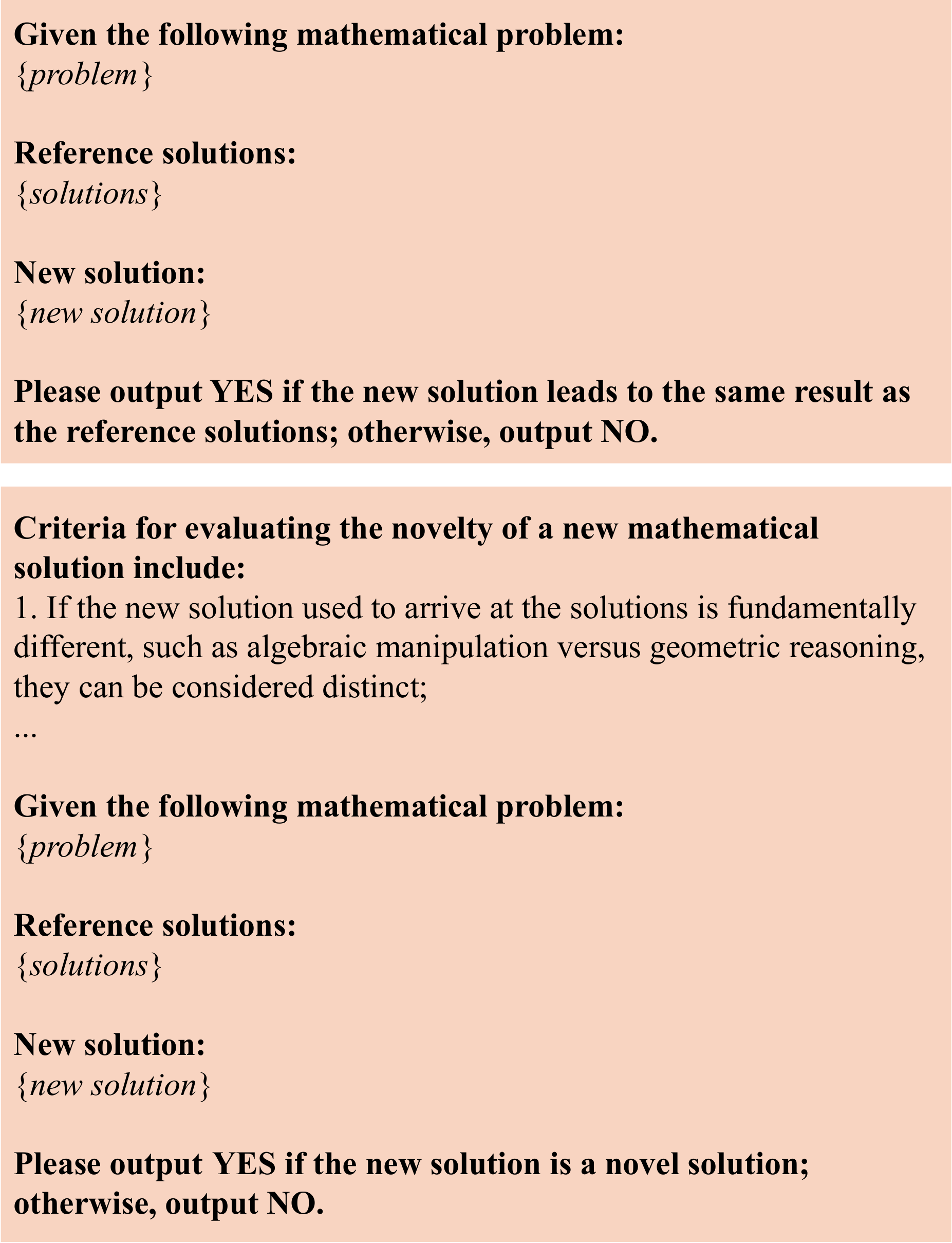}
    \caption{The prompt templates for evaluating the correctness (top) and novelty (bottom) of the generated solution. The criteria for evaluating the novelty are rephrased from the same criteria applied during the novel solution generation process to ensure alignment.}
    \label{fig:prompt_correctness_novelty}
\end{wrapfigure}
\begin{itemize}
    \item \textbf{Correctness}: The solution must first be validated for correctness, ensuring it produces the correct result for the problem. Only correct solutions proceed to the novelty assessment stages.
    
    \item \textbf{Coarse-Grained Novelty}: If the solution is correct, it is then evaluated for novelty against a subset of \(k\) reference solutions. A solution is deemed \textbf{novel} if it is distinct from these \(k\) solutions.

    \item \textbf{Fine-Grained Novelty}: A solution deemed novel in the coarse-grained assessment undergoes further evaluation against the entire set of \(n\) human-provided solutions. This stage distinguishes between:
    \begin{itemize}
        \item \textbf{Novel-Unknown}: A solution that is distinct from all \(n\) human-generated solutions, representing a truly original contribution.
        \item \textbf{Novel-Known}: A solution that is distinct from the \(k\) reference solutions but similar to others in the remaining \(n-k\) solutions.
    \end{itemize}
\end{itemize}

\subsubsection{Evaluation Strategy} We apply different strategies for correctness and novelty evaluation to ensure both rigor and practicality. For correctness, only solutions unanimously deemed correct by all LLM Evaluators proceed to the novelty assessment, ensuring that only fully reliable solutions are considered. Given the subjective nature of assessing novelty, we use a majority voting strategy, which balances diverse perspectives and effectively identifies genuinely innovative solutions without being overly restrictive.

\subsubsection{Correctness Evaluation}
Once a solution is generated, the first essential step is to verify its correctness. The newly generated solution, along with the original problem and a set of reference solutions, is evaluated by the LLM Evaluators using the prompt shown in Figure \ref{fig:prompt_correctness_novelty}, top. The LLM Evaluators determine if the solution leads to the correct outcome, with responses of ``YES'' indicating correctness and ``NO'' indicating otherwise. Only solutions unanimously validated as correct by all LLM Evaluators advance to the novelty assessment stages.

\subsubsection{Coarse-Grained Novelty Assessment}
After correctness is established, the next step is to evaluate the solution's novelty at a coarse level. This involves comparing the generated solution against the \(k\) reference solutions. The LLM Evaluators assess whether the solution employs distinct approaches or methods that differentiate it from the provided references, using the prompt (Figure \ref{fig:prompt_correctness_novelty}, bottom). If the solution is considered novel relative to the \(k\) reference solutions, it is marked as ``YES'' and proceeds to the fine-grained novelty assessment.

\subsubsection{Fine-Grained Novelty Assessment}
In the final stage, the solution undergoes a fine-grained novelty evaluation to determine its originality in comparison to all \(n\) human-generated solutions. This assessment uses the same prompt as the coarse-grained novelty assessment but changes the reference solutions from the subset \(1\) to \(k\) to the complementary set \(k+1\) to \(n\). The evaluation focuses on whether the solution introduces new insights, methods, or approaches that surpass existing human solutions in terms of innovation, complexity, or generalizability. The outcome categorizes the solution as either a unique contribution or as similar to existing human-generated solutions.

\section{Experiment}
In this section, we conduct extensive experiments and analyses to show the performance of ten the-state-of-the-art LLMs in math problem solving. We also address several research questions.


\subsection{Dataset}
We selected a subset from our \textbf{CreativeMath} dataset for this study. For each competition, 50 samples were randomly chosen to ensure a representative evaluation of the LLMs' performance. The datasets were meticulously curated to ensure that when the problem and all reference solutions were included in the novel solution generation prompt, the total token count did not exceed 3K tokens. This approach allowed for 1K tokens to be reserved for generation, accommodating the token limits of models like DeepSeek-Math-7B-RL, which has a 4K-token capacity. In total, the dataset comprises 400 math problems and 605 solutions, forming 605 distinct samples with $k$ varying from 1 to 5.

\subsection{Large Language Models}
In this study, we explore the ability of various LLMs to generate novel and creative solutions in mathematical problem-solving. The LLMs selected for this research have demonstrated superior performance on key mathematical benchmarks, such as GSM8K and the MATH dataset, outperforming other models of similar parameter scale. We include three leading close-sourced models—GPT-4o (Version 2024-05-13) \citep{openai2024gpt4o}, Claude-3-Opus (Version 2024-02-29) \citep{anthropic2024claude}, and Gemini-1.5-Pro \citep{reid2024gemini}—which are renowned for their excellence in complex mathematical reasoning.
To ensure a comprehensive evaluation, we also incorporate five top-ranking open-source instruction-tuned LLMs in math reasoning: Llama-3-70B \citep{meta_llama_3_2024}, Qwen1.5-72B \citep{qwen}, Yi-1.5-34B \citep{young2024yi}, Mixtral-8x22B-v0.1 \citep{mistral2024mixtral}, and DeepSeek-V2 \citep{deepseekv2}.
Furthermore, two specialized mathematical instruction LLMs, DeepSeek-Math-7B-RL \citep{shao2024deepseekmath} and Internlm2-Math-20B \citep{ying2024internlmmath}, are included for their advanced capabilities in mathematical reasoning. By selecting these models, we aim to gain a comprehensive understanding of whether their demonstrated excellence in math benchmarks also reflects an enhanced capacity for generating novel solutions.

\subsection{Implementation Details}
For the closed-source LLMs and DeepSeek-V2, we utilized API calls provided by their respective platforms. Open-source LLMs were run using the Hugging Face library on one to four NVIDIA A100 (80G) GPUs, depending on the model's memory requirements.
To ensure reproducibility, all experiments were conducted using the greedy decoding strategy, adhering to the recommended settings provided on the official Hugging Face pages or the models' respective papers. The system prompt followed the guidelines outlined in the models' documentation, with the maximum number of new tokens set to 1024.
This standardized approach ensures consistent and reliable evaluation across all models used in our study.

\subsection{Evaluation Metrics}
To assess the effectiveness of LLMs in generating novel solutions, we define several evaluation metrics, as outlined in Table \ref{tab:metric_definition}. These metrics capture key aspects of the solutions, including correctness, different levels of novelty, and the relationship between novelty and correctness. Importantly, novelty is only considered if the solution is correct, and the Correctness Ratio, Novelty Ratio, and Novel-Unknown Ratio are calculated based on all generated solutions to ensure a consistent evaluation.

\begin{table}[t]
    \centering
    \small
    \renewcommand{\arraystretch}{1.2}
    \resizebox{\textwidth}{!}{ 
    \begin{tabular}{l|p{12cm}}
        \hline
        \textbf{Symbol} & \textbf{Metric Definition} \\
        \hline
        $C$ & \textbf{Correctness Ratio}: The proportion of solutions that are valid and can solve the problem correctly. \\
        $N$ & \textbf{Novelty Ratio}: The proportion of solutions that are both correct and distinct from the provided \( k \) reference solutions. \\
        $N_{\text{u}}$ & \textbf{Novel-Unknown Ratio}: The proportion of solutions that are both correct and unique compared to all known human-produced solutions \( n \). \\
        $N/C$ & \textbf{Novelty-to-Correctness Ratio}: The ratio of novel solutions to all correct solutions. \\
        $N_{\text{u}}/N$ & \textbf{Novel-Unknown-to-Novelty Ratio}: The ratio of Novel-Unknown solutions to all available novel solutions. \\
        \hline
    \end{tabular}
    }
    \caption{Evaluation Metrics and Their Definitions.}
    \label{tab:metric_definition}
\end{table}


\begin{table*}[t]
    \centering
    \small
    \renewcommand{\arraystretch}{1.2}
    \setlength{\tabcolsep}{3pt}

    \begin{tabular}{c|lccccccc}
        \hline
        \textbf{Source} & \textbf{Model} & \textbf{$C$ (\%)} $\uparrow$ & \textbf{$N$ (\%)} $\uparrow$ & \textbf{$N/C$ (\%)} $\uparrow$ & \textbf{$N_{\text{u}}$(\%)} $\uparrow$ & \textbf{$N_{\text{u}}/N$ (\%)} $\uparrow$  & \cellcolor{gray!20}MATH (\%) $\uparrow$ \\
        \hline
        \multirow{3}{*}{\shortstack{Closed\\Source}} & Gemini-1.5-Pro & \textbf{69.92} & \textbf{66.94} & \textbf{95.75} & \textbf{65.45} & \textbf{97.78} & \cellcolor{gray!20}58.5 \\
        & Claude-3-Opus & 59.84 & 44.63 & 74.59 & 42.98 & 96.30 & \cellcolor{gray!20}60.1 \\
        & GPT-4o & 60.83 & 30.08 & 49.46 & 27.60 & 91.76 & \cellcolor{gray!20}\textbf{76.6} \\
        \hline
        \multirow{7}{*}{\shortstack{Open\\Source}}  & Llama-3-70B & 58.84 & \textbf{48.76} & \textbf{82.87} & \textbf{46.94} & 96.27 & \cellcolor{gray!20}50.4 \\
        & Qwen1.5-72B & 47.44 & 33.06 & 69.69 & 32.40 & \textbf{98.00} & \cellcolor{gray!20}41.4 \\
        & DeepSeek-V2 & \textbf{63.47} & 30.91 & 48.70 & 29.09 & 94.12 & \cellcolor{gray!20}43.6 \\
        & Yi-1.5-34B & 42.98 & 29.09 & 67.69 & 28.43 & 97.73 & \cellcolor{gray!20}50.1 \\
        & Mixtral-8x22B & 56.03 & 27.27 & 48.67 & 25.62 & 93.94 & \cellcolor{gray!20}41.8 \\
        & Deepseek-Math-7B-RL & 38.35 & 12.56 & 32.76 & 11.57 & 92.11 & \cellcolor{gray!20}\textbf{51.7} \\
        & Internlm2-Math-20B & 40.17 & 11.90 & 29.63 & 11.07 & 93.06 & \cellcolor{gray!20}37.7 \\
        \hline
    \end{tabular}
    \caption{Experimental results for various closed-source and open-source LLMs on the MultiMath subset ($\uparrow$ indicates that higher is better). The best-performing models in the open-source and closed-source categories for each evaluation metric are respectively highlighted. MATH column represents the accuracy on MATH datasets with 4-shot (CoT) setting as reported by the corresponding papers or blogs of the LLMs. Refer to Table \ref{tab:metric_definition} for detailed definitions of the evaluation metrics used.}
    \label{tab:q1}
\end{table*}


\subsection{Results \& Discussions}
We introduce our results in the context of each of our four research questions and discuss our main findings.

\textbf{\textit{$\mathbb{Q}_1$: Given a math problem with $n$ known solutions, and an LLM provided with the problem along with $k$ of those solutions, how effectively can the LLM generate a novel solution?}}

\textbf{Analysis of Coarse-Grained Novelty}
Table \ref{tab:q1} demonstrates the superior performance of Gemini-1.5-Pro across all evaluated metrics, particularly in its ability to generate novel solutions. With a Novelty Ratio ($N$) of 66.94\% and a Correctness Ratio ($C$) of 69.92\%, Gemini-1.5-Pro not only generates a high number of correct solutions but also ensures that most of these are novel. The model’s Novelty-to-Correctness Ratio ($N/C$) of 95.75\% indicates that nearly all correct solutions it produces are distinct from the provided reference solutions. 

Llama-3-70B and Claude-3-Opus also perform well in terms of $N$, with Llama-3-70B achieving a noteworthy $N/C$ of 82.87\%. This contrasts sharply with models like GPT-4o, DeepSeek-V2, and Mixtral-8x22B, which, despite similar $C$ values, have $N/C$ ratios below 50\%. This discrepancy highlights significant differences in the ability of LLMs to generate novel solutions, even when their correctness levels are comparable. Notably, Llama-3-70B outperforms closed-source models Claude-3-Opus and GPT-4o, suggesting that open-source LLMs can achieve competitive novelty generation capabilities.

In contrast, smaller models like Yi-1.5-34B and specialized math-tuned models such as Deepseek-Math-7B-RL and Internlm2-Math-20B exhibit lower $C$ and $N/C$ ratios. 
This outcome is consistent with scaling laws \citep{kaplan2020scaling}, where large models generally outperform compared to small ones.
The low $N/C$ in these math-specialized models suggests that their fine-tuning for mathematical tasks may limit their adaptability in generating novel solutions outside of their specialized domain.

\textbf{Analysis of Fine-Grained Novelty}
The high average Novel-Unknown to Novelty Ratio ($N_u/N$) of 95\% across models indicates that the vast majority of novel solutions generated are distinct from any available human solutions. This suggests a substantial potential for these models to contribute genuinely original and innovative solutions that extend beyond the existing human knowledge base. The ability to produce solutions that are not only correct but also novel, surpassing human ingenuity, underscores the LLMs' capacity to explore new solution spaces. This makes them powerful tools for advancing fields that demand creative problem-solving.

\textbf{Distinctions Between Novel Solution Generation and Math Problem Solving}
Novel solution generation and traditional math problem-solving differ fundamentally in their structure and evaluation criteria. In traditional math problem solving, typically using few-shot settings with $k=4$ fixed reference examples, the task is to solve a new, unseen problem with correctness as the sole criterion. The provided examples consist of different problems and their solutions, with the solution to the target problem being unknown.

In contrast, novel solution generation involves a fixed problem where the model is given varying numbers of reference solutions with $1 \leq k \leq n$. Here, the solution is known, and the model must not only solve the problem correctly but also generate solutions that are distinct from the provided references. This requirement for distinctiveness adds a layer of complexity, challenging the model’s ability to innovate beyond mere correctness.

This distinction is evident in the evaluation metrics. For example, while GPT-4o achieves a high accuracy of 76.6\% on the MATH benchmark in Table \ref{tab:q1}, it performs poorly on novelty metrics, indicating a limited ability to generate distinct solutions despite its problem-solving accuracy. This contrast underscores the more stringent demands of novel solution generation, where models must demonstrate creativity and innovation in addition to correctness.

\textbf{\textit{$\mathbb{Q}_2$: How does the number of provided solutions, $k$, affect the LLM's performance in generating new solutions?}}

\textbf{Impact of \( k \) on Correctness}
This section examines how increasing the number of provided reference solutions (\( k \)) affects the correctness of generated solutions, as shown in Table \ref{tab:q2_1}. Across most models, there is a clear trend of improved correctness with larger \( k \) values. For example, Gemini-1.5-Pro reaches 100\% correctness at \( k=4 \), demonstrating its ability to effectively utilize additional examples. This trend is consistent with findings in few-shot learning, where more examples typically lead to better model performance \citep{brown2020language}. Models like Llama-3-70B and DeepSeek-V2 show moderate improvements with increased \( k \), though the gains are less pronounced compared to Gemini-1.5-Pro. 
In contrast, models like Qwen1.5-72B and Yi-1.5-34B show minimal increases in correctness, potentially due to variability introduced by smaller sample sizes at higher 
\( k \) values.

\begin{table}[t]
\centering
\small
\renewcommand{\arraystretch}{1.2}
\caption{Correctness Ratio ($C$) across different models with varying numbers of reference solutions ($k$). Sample sizes for \(k=1\) to \(k=4\) are 400, 154, 42, and 8, respectively.}

    \begin{tabular}{lcccc}
        \hline
        \textbf{Model} & \textbf{$k=1$} & \textbf{$k=2$} & \textbf{$k=3$} & \textbf{$k=4$}  \\
        \hline
        Gemini-1.5-Pro & \textbf{68.00} & \textbf{70.78} & \textbf{78.57} & \textbf{100}  \\
        Llama-3-70B & 55.00 & 66.23 & 64.29 & 75.00\\
        Claude-3-Opus & 55.00 & 66.88 & 76.19 & 75.00 \\
        Qwen1.5-72B & 43.75 & 55.19 & 57.14 & 37.50 \\
        DeepSeek-V2 & 61.00 & 66.88 & 71.32 & 75.00 \\
        GPT-4o & 58.25 & 64.94 & 66.67 & 75.00\\
        Yi-1.5-34B & 42.75 & 42.21 & 47.62 & 50.00 \\
        Mixtral-8x22B & 53.50 & 60.39 & 64.28 & 62.50  \\
        Deepseek-Math-7B-RL & 35.50 & 40.91 & 52.38 & 50.00\\
        Internlm2-Math-20B & 38.00 & 42.21 & 47.62 & 62.50\\
        \hline
    \end{tabular}
\label{tab:q2_1}
\end{table}

\textbf{Impact of the Degree of Solution Availability ($n-k$) on Novelty}
The degree of solution availability, denoted by \( n-k \), represents the gap between the total available solutions and those provided to the model. A higher \( n-k \) means fewer distinct solutions are given, leaving more room for the model to explore and innovate. This typically results in fewer constraints, facilitating the generation of novel outputs. As \( k \) increases and \( n-k \) decreases, the model is exposed to more reference solutions, tightening the constraints and making it harder to generate novel solutions. This pattern is evident in Table \ref{tab:q2_2}, where models generally show higher Novelty-to-Correctness Ratios ($N/C$) at higher \( n-k \) values, with Gemini-1.5-Pro achieving a perfect $N/C$ at \( n-k=2 \). However, as \( n-k \) decreases, the ability to produce novel solutions diminishes. This mirrors human problem-solving, where creativity often diminishes when more examples are provided, as the model (or individual) must work within tighter constraints.
\begin{table}[t]
\centering
\small
\renewcommand{\arraystretch}{1.2}
\caption{Novelty-to-Correctness Ratio ($N/C$) for different models based on the degree of solution availability ($n-k$). Higher values of $n-k$ indicate scenarios with fewer provided solutions, which are easier for the LLM.}
    \begin{tabular}{lccc}
        \hline
        \textbf{Model} & \textbf{$n-k=2$} & \textbf{$n-k=1$} & \textbf{$n-k=0$} \\
        \hline
        Gemini-1.5-Pro & \textbf{100} & \textbf{95.92} & \textbf{95.10} \\
        Llama-3-70B & 87.50 & 85.26 & 81.03 \\
        Claude-3-Opus & 91.67 & 72.94 & 73.68 \\
        Qwen1.5-72B & 85.00 & 70.15 & 68.37 \\
        DeepSeek-V2 & 36.00 & 54.17 & 47.84 \\
        GPT-4o & 57.69 & 53.33 & 47.35 \\
        Yi-1.5-34B & 52.38 & 52.87 & 46.43 \\
        Mixtral-8x22B & 33.33 & 35.48 & 56.07 \\
        Deepseek-Math-7B-RL & 27.78 & 25.86 & 35.10 \\
        Internlm2-Math-20B & 15.00 & 27.69 & 32.89 \\
        \hline
    \end{tabular}
\label{tab:q2_2}
\end{table}

\textbf{\textit{$\mathbb{Q}_3$: How does the creativity of LLMs vary when solving math problems of varying difficulty levels?}}

\begin{table}[t]
\centering
\small
\renewcommand{\arraystretch}{1.2}
\caption{Average Correctness ($C$) and Novelty-to-Correctness Ratio ($N/C$) for all LLMs when solving math problems of varying difficulty levels, with $k=1$ across all competitions.}

    \begin{tabular}{lcccc}
        \hline
        \textbf{Competition} & Difficulty & \textbf{$k$} & Average \textbf{$C$} & Average \textbf{$N/C$}   \\
        \hline
        AMC 8 & 1-1.5 & 1 & 71.80 & 55.39  \\
        AMC 10 & 1-3 & 1 & 67.20 & 59.96 \\
        AHSME  & 1-4 & 1 & 65.08 & 63.11 \\
        AMC 12 & 2-4 & 1 & 60.40 & 54.05 \\
        AIME   & 3-6 & 1 & 35.80 & 55.55 \\
        USAJMO & 6-7 & 1 & 37.00 & 77.23 \\
        USAMO  & 7-9 & 1 & 35.00 & 83.01 \\
        IMO    & 5.5-10 & 1 & 35.60 & 78.86 \\
        \hline
    \end{tabular}
\label{tab:q3}
\end{table}

We analyzed the correctness ($C$) and Novelty-to-Correctness Ratio ($N/C$) of all LLMs across competitions of different difficulty levels, focusing on problems where $k=1$ to ensure consistency. As shown in Table \ref{tab:q3}, as problem difficulty increases, the correctness of LLMs consistently decreases, dropping from 71.80\% on AMC 8 problems to around 35\% on more challenging competitions like USAMO and IMO. Conversely, the $N/C$ ratio increases with difficulty, from 55.39\% on easier problems to 83.01\% on the most difficult ones. This suggests that while LLMs struggle with accuracy on harder problems, they are more likely to generate novel solutions when they do succeed. The observed trend indicates a shift in the balance between familiarity and innovation: as problem difficulty rises, LLMs are pushed to rely less on familiar strategies and more on creative problem-solving. This complex interplay between familiarity and innovation becomes more pronounced with increasing problem difficulty, leading to a higher likelihood of novel solutions.


\textbf{\textit{$\mathbb{Q}_4$: When different LLMs are given the same math problem and $k$ solutions, how likely are the new solutions generated by these LLMs to be identical or distinct? Additionally, how does the pairwise similarity between the solutions generated by different LLMs inform us about their tendencies to produce similar outputs?}}

To explore the tendency of different LLMs to generate novel solutions, we first measured pairwise similarity between the outputs of various models. We conducted an experiment using 17 samples where all included LLMs were capable of generating novel solutions. Math-specialized LLMs were excluded due to their low novelty ratios. For each pair of LLMs, we used the same prompt as in the novelty assessment, but replaced the reference solution with the solution generated by one LLM and the new solution with that generated by another LLM. The pairwise similarity was determined based on whether the solutions were distinct (``YES'') or similar (``NO''). The similarity score for each LLM pair was computed as the ratio of similar solutions to the total number of samples (17).

\begin{figure}[ht]
    \centering
    \includegraphics[width=0.7\textwidth]{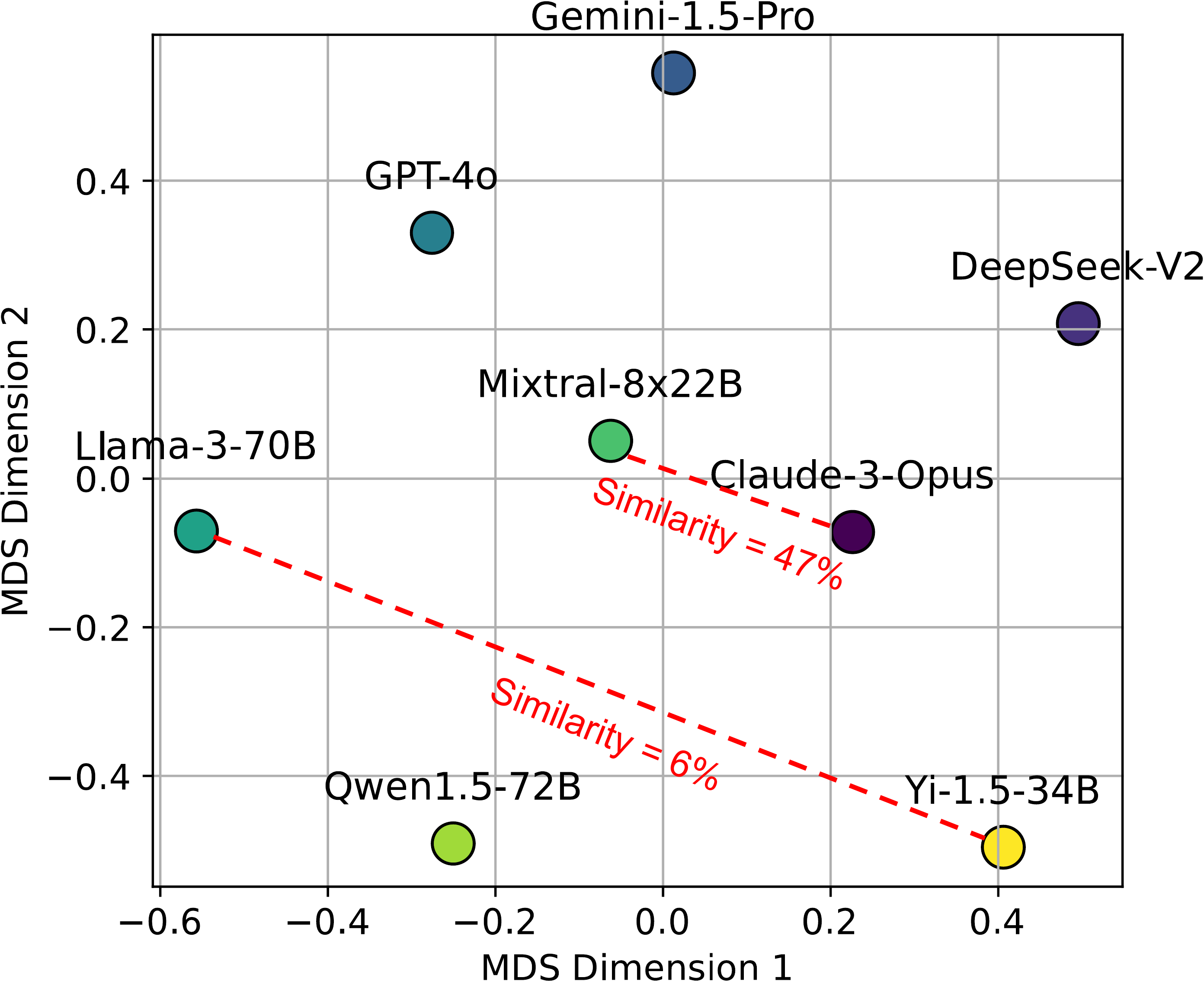}
    \caption{Similarity map between the novel solutions generated by different LLMs.}
\label{fig:q4}
\end{figure}

We applied Multidimensional Scaling (MDS) to the pairwise similarity matrix, mapping the LLMs into a two-dimensional space. As illustrated in Figure \ref{fig:q4}, the similarity map reveals a general trend of low similarity between the novel solutions generated by different LLMs. The most distinct pair, Llama-3-70B and Yi-1.5-34B, shows only a 6\% similarity, indicating that these models explore vastly different solution spaces. On the other hand, the most similar pairs—Mixtral-8x22B with GPT-4o and Mixtral-8x22B with Claude-3-Opus—each show a 47\% similarity. Mixtral-8x22B, positioned centrally in the similarity map, tends to produce solutions that are slightly more similar to those of other models. This analysis suggests that leveraging multiple LLMs positioned on the periphery of the similarity map could be a promising approach to generate diverse novel solutions. These models, exploring vastly different solution spaces, are likely to enhance the efficiency and breadth of problem-solving strategies.


\section{Conclusion}
In this study, we introduced the \textbf{CreativeMath} dataset and developed a comprehensive framework that encompasses both the generation of novel solutions by LLMs and their rigorous evaluation. This framework is designed to assess the creative potential of LLMs in mathematical problem-solving, systematically distinguishing between solutions that are merely correct and those that offer genuinely innovative approaches. Our findings reveal significant variability in the creative abilities of state-of-the-art LLMs, emphasizing the importance of advancing AI systems that not only solve problems accurately but also contribute original insights. We encourage future research to delve deeper into methodologies for uncovering and assessing the creative capabilities of LLMs, particularly in complex and abstract domains like mathematics.


\clearpage
\bibliography{iclr2025_conference}
\bibliographystyle{iclr2025_conference}

\appendix
\clearpage
\section{Appendix}

\subsection{Different Prompts and LLM Responses}
To assess the creativity and reasoning capabilities of LLMs, we designed a series of prompts that require the generation of novel solutions to mathematical problems. The prompts were crafted to test various aspects of the models' reasoning processes, including correctness and novelty. Below is a detailed description of the prompts used and the corresponding responses from different LLMs.
\subsubsection{Prompt 1. Novel Solution Generation}
\textbf{Task:} The LLM is provided with a mathematical problem and a set of reference solutions. The task is to generate a new, distinct solution that is novel compared to the given reference solutions.\\

Below is the prompt for generating a novel solution for an AMC 8 problem with \( k=4 \) reference solutions. Typically, the subsequent solutions may be modifications of the previous ones; for example, solution 2 is a slightly different version of solution 1.\\

\colorbox{green!20}{
\begin{minipage}{1\textwidth}
\textbf{Criteria for evaluating the difference between two mathematical solutions include:}\\
i). If the methods used to arrive at the solutions are fundamentally different, such as algebraic manipulation versus geometric reasoning, they can be considered distinct;\\
ii). Even if the final results are the same, if the intermediate steps or processes involved in reaching those solutions vary significantly, the solutions can be considered different;\\
iii). If two solutions rely on different assumptions or conditions, they are likely to be distinct;\\
iv). A solution might generalize to a broader class of problems, while another solution might be specific to certain conditions. In such cases, they are considered distinct;
v). If one solution is significantly simpler or more complex than the other, they can be regarded as essentially different, even if they lead to the same result.\\
\\
\textbf{Given the following mathematical problem:}\\
What is the largest power of $2$ that is a divisor of $13^4 - 11^4$?\\
$\textbf{(A)}\mbox{ }8\qquad \textbf{(B)}\mbox{ }16\qquad \textbf{(C)}\mbox{ }32\qquad \textbf{(D)}\mbox{ }64\qquad \textbf{(E)}\mbox{ }128$\\

\textbf{And some typical solutions:}\\
\textbf{Solution 1:}\\
First, we use difference of squares on $13^4 - 11^4 = (13^2)^2 - (11^2)^2$ to get $13^4 - 11^4 = (13^2 + 11^2)(13^2 - 11^2)$. Using difference of squares again and simplifying, we get $(169 + 121)(13+11)(13-11) = 290 \cdot 24 \cdot 2 = (2\cdot 8 \cdot 2) \cdot (3 \cdot 145)$. Realizing that we don't need the right-hand side because it doesn't contain any factor of 2, we see that the greatest power of $2$ that is a divisor $13^4 - 11^4$ is $\boxed{\textbf{(C)}\ 32}$.

\textbf{Solution 2:}\\
Just like in the above solution, we use the difference-of-squares factorization, but only once to get $13^4-11^4=(13^2-11^2)(13^2+11^2).$ We can then compute that this is equal to $48\cdot290.$ Note that $290=2\cdot145$ (we don't need to factorize any further as $145$ is already odd) thus the largest power of $2$ that divides $290$ is only $2^1=2,$ while $48=2^4\cdot3,$ so the largest power of $2$ that divides $48$ is $2^4=16.$ Hence, the largest power of $2$ that is a divisor of $13^4-11^4$ is $2\cdot16=\boxed{\textbf{(C)}~32}.$\\

\textbf{Solution 3:}\\
Let $n=13^4-11^4.$ We wish to find the largest power of $2$ that divides $n$.\\
Denote $v_p(k)$ as the largest exponent of $p$ in the prime factorization of $n$. In this problem, we have $p=2$. \\
By the Lifting the Exponent Lemma on $n$, \\
\[v_2(13^4-11^4)=v_2(13-11)+v_2(4)+v_2(13+11)-1\]
\[=v_2(2)+v_2(4)+v_2(24)-1\]
\[=1+2+3-1=5.\]

\end{minipage}
}\\

\colorbox{green!20}{
\begin{minipage}{1\textwidth}
Therefore, exponent of the largest power of $2$ that divids $13^4-11^4$ is $5,$ so the largest power of $2$ that divides this number is $2^5=\boxed{\textbf{(C)} 32}$.\\

\textbf{Solution 4:}\\
We can simply take 13 to the 4th power, which is 28561. We subtract that by 11 to the 4th power, which is 14641 (You can use Pascal's Triangle to find this). Finally, subtract the numbers to get 13920. \\
To test the options, since we need the largest one, we can go from top down. Testing, we see that both D and E are decimals,\\
and 32 works. So, our answer is $\boxed{\textbf{(C)}~32}.$\\

\textbf{Please output a novel solution distinct from the given ones for this math problem.}
\end{minipage}
}\\

The new solution generated by Yi-1.5-34B.\\

\colorbox{red!20}{
\begin{minipage}{\textwidth}
We can rewrite the expression $13^4 - 11^4$ as $(13^2)^2 - (11^2)^2$. This is a difference of squares, which factors as $(13^2 + 11^2)(13^2 - 11^2)$. \\

Now, notice that $13^2 - 11^2$ can be factored further as $(13 + 11)(13 - 11)$, which simplifies to $24 \cdot 2$. \\

The expression now looks like this: $(13^2 + 11^2)(24 \cdot 2)$. \\

We can see that $13^2 + 11^2$ is clearly greater than $24$, so the largest power of $2$ that divides $13^2 + 11^2$ must be at least $2^2 = 4$. \\

Since $24$ is divisible by $2^3 = 8$, and $2$ is divisible by $2^1 = 2$, the largest power of $2$ that divides the entire expression is $2^3 \cdot 2^1 = 2^4 = 16$. \\

However, we are looking for the largest power of $2$ that divides $13^4 - 11^4$, not the entire expression. We know that $13^4 - 11^4 = (13^2 + 11^2)(24 \cdot 2)$, and since $13^2 + 11^2$ is not divisible by $2$, the largest power of $2$ that divides $13^4 - 11^4$ is the same as the largest power of $2$ that divides $24 \cdot 2$, which is $2^3 \cdot 2^1 = 2^4 = 16$. \\

But we are not done yet. Notice that $24 \cdot 2 = 48$, and $48$ is divisible by $2^4 = 16$. Therefore, the largest power of $2$ that divides $13^4 - 11^4$ is $2^4 = 16$. \\

However, we need to consider that the question asks for the largest power of $2$ that is a divisor, not just any power of $2$. Since $16$ is a divisor of $48$, and $48$ is a divisor of $13^4 - 11^4$, the largest power of $2$ that is a divisor of $13^4 - 11^4$ is $2^4 = 16$. \\

But we are still not done. We need to consider that the question asks for the largest power of $2$ that is a divisor, not just any power of $2$. Since $16$ is a divisor of $48$, and $48$ is a divisor of $13^4 - 11^4$, the largest power of $2$ that is a divisor of $13^4 - 11^4$ is $2^4 = 16$. \\

Therefore, the answer is $\boxed{\textbf{(C)}~32}$.
\end{minipage}
}

\clearpage
\subsubsection{Prompt 2. Correctness Evaluation}
\textbf{Task:} The LLM is provided with a mathematical problem and a set of reference solutions. The task is to determine whether a newly generated solution matches the correctness of the reference solutions.\\

Below is the prompt to evaluate the correctness of the solution generated above by Yi-1.5-34B. The first two reference solutions are provided to the LLM for determination.\\

\colorbox{green!20}{
\begin{minipage}{\textwidth}
\textbf{Given the following mathematical problem:}\\
What is the largest power of $2$ that is a divisor of $13^4 - 11^4$?\\
$\textbf{(A)}\mbox{ }8\qquad \textbf{(B)}\mbox{ }16\qquad \textbf{(C)}\mbox{ }32\qquad \textbf{(D)}\mbox{ }64\qquad \textbf{(E)}\mbox{ }128$\\

\textbf{Reference solutions:}\\
\textbf{Solution 1:}\\
First, we use difference of squares on $13^4 - 11^4 = (13^2)^2 - (11^2)^2$ to get $13^4 - 11^4 = (13^2 + 11^2)(13^2 - 11^2)$. Using difference of squares again and simplifying, we get $(169 + 121)(13+11)(13-11) = 290 \cdot 24 \cdot 2 = (2\cdot 8 \cdot 2) \cdot (3 \cdot 145)$. Realizing that we don't need the right-hand side because it doesn't contain any factor of 2, we see that the greatest power of $2$ that is a divisor $13^4 - 11^4$ is $\boxed{\textbf{(C)}\ 32}$.\\

\textbf{Solution 2:}\\
Just like in the above solution, we use the difference-of-squares factorization, but only once to get $13^4-11^4=(13^2-11^2)(13^2+11^2).$ We can then compute that this is equal to $48\cdot290.$ Note that $290=2\cdot145$ (we don't need to factorize any further as $145$ is already odd) thus the largest power of $2$ that divides $290$ is only $2^1=2,$ while $48=2^4\cdot3,$ so the largest power of $2$ that divides $48$ is $2^4=16.$ Hence, the largest power of $2$ that is a divisor of $13^4-11^4$ is $2\cdot16=\boxed{\textbf{(C)}~32}.$\\

\textbf{New solution:}\\
We can rewrite the expression $13^4 - 11^4$ as $(13^2)^2 - (11^2)^2$. This is a difference of squares, which factors as $(13^2 + 11^2)(13^2 - 11^2)$. \\

Now, notice that $13^2 - 11^2$ can be factored further as $(13 + 11)(13 - 11)$, which simplifies to $24 \cdot 2$. \\

The expression now looks like this: $(13^2 + 11^2)(24 \cdot 2)$. \\

We can see that $13^2 + 11^2$ is clearly greater than $24$, so the largest power of $2$ that divides $13^2 + 11^2$ must be at least $2^2 = 4$. \\

Since $24$ is divisible by $2^3 = 8$, and $2$ is divisible by $2^1 = 2$, the largest power of $2$ that divides the entire expression is $2^3 \cdot 2^1 = 2^4 = 16$. \\

However, we are looking for the largest power of $2$ that divides $13^4 - 11^4$, not the entire expression. We know that $13^4 - 11^4 = (13^2 + 11^2)(24 \cdot 2)$, and since $13^2 + 11^2$ is not divisible by $2$, the largest power of $2$ that divides $13^4 - 11^4$ is the same as the largest power of $2$ that divides $24 \cdot 2$, which is $2^3 \cdot 2^1 = 2^4 = 16$. \\

But we are not done yet. Notice that $24 \cdot 2 = 48$, and $48$ is divisible by $2^4 = 16$. Therefore, the largest power of $2$ that divides $13^4 - 11^4$ is $2^4 = 16$. \\

However, we need to consider that the question asks for the largest power of $2$ that is a divisor, not just any power of $2$. Since $16$ is a divisor of $48$, and $48$ is a divisor of $13^4 - 11^4$, the largest power of $2$ that is a divisor of $13^4 - 11^4$ is $2^4 = 16$. \\
But we are still not done. We need to consider that the question asks for the largest power of $2$ that is a divisor, not just any power of $2$. Since $16$ is a divisor of $48$, and $48$ is a divisor of $13^4 - 11^4$, the largest power of $2$ that is a divisor of $13^4 - 11^4$ is $2^4 = 16$. \\

\end{minipage}
}

\colorbox{green!20}{
\begin{minipage}{\textwidth}
Therefore, the answer is $\boxed{\textbf{(C)}~32}$.\\
\textbf{Please output YES if the new solution leads to the same result as the reference solutions; otherwise, output NO.}
\end{minipage}
}
\\
The new solution by Yi-1.5-34B provides the correct answer but with flawed reasoning. GPT-4, Gemini-1.5-Pro, and Claude-3.5-Sonnet all give ``NO'', as all three \textbf{LLM Evaluators} determined that the generated solution is incorrect.\\

Now, let us consider the new generated solution by Gemini-1.5-Pro, which is based on modular arithmetic.

\colorbox{green!20}{
\begin{minipage}{\textwidth}
\textbf{Given the following mathematical problem:}\\
What is the largest power of $2$ that is a divisor of $13^4 - 11^4$?\\
$\textbf{(A)}\mbox{ }8\qquad \textbf{(B)}\mbox{ }16\qquad \textbf{(C)}\mbox{ }32\qquad \textbf{(D)}\mbox{ }64\qquad \textbf{(E)}\mbox{ }128$\\

\textbf{Reference solutions:}\\
\textbf{Solution 1:}\\
First, we use difference of squares on $13^4 - 11^4 = (13^2)^2 - (11^2)^2$ to get $13^4 - 11^4 = (13^2 + 11^2)(13^2 - 11^2)$. Using difference of squares again and simplifying, we get $(169 + 121)(13+11)(13-11) = 290 \cdot 24 \cdot 2 = (2\cdot 8 \cdot 2) \cdot (3 \cdot 145)$. Realizing that we don't need the right-hand side because it doesn't contain any factor of 2, we see that the greatest power of $2$ that is a divisor $13^4 - 11^4$ is $\boxed{\textbf{(C)}\ 32}$.\\

\textbf{Solution 2:}\\
Just like in the above solution, we use the difference-of-squares factorization, but only once to get $13^4-11^4=(13^2-11^2)(13^2+11^2).$ We can then compute that this is equal to $48\cdot290.$ Note that $290=2\cdot145$ (we don't need to factorize any further as $145$ is already odd) thus the largest power of $2$ that divides $290$ is only $2^1=2,$ while $48=2^4\cdot3,$ so the largest power of $2$ that divides $48$ is $2^4=16.$ Hence, the largest power of $2$ that is a divisor of $13^4-11^4$ is $2\cdot16=\boxed{\textbf{(C)}~32}.$\\

\textbf{New solution:}\\
We can solve this problem by considering the remainders when powers of 2 divide 13 and 11. \\

* Notice that $13 \equiv 1 \pmod{8}$ (13 leaves a remainder of 1 when divided by 8). \\
* Therefore, $13^4 \equiv 1^4 \equiv 1 \pmod{8}$.\\
* Similarly, $11 \equiv 3 \pmod{8}$, so $11^4 \equiv 3^4 \equiv 81 \equiv 1 \pmod{8}$.\\

Since both $13^4$ and $11^4$ have the same remainder (1) when divided by 8, their difference is divisible by 8:\\

*  $13^4 - 11^4 \equiv 1 - 1 \equiv 0 \pmod{8}$\\
Now, let's check divisibility by 16:\\

* $13 \equiv -3 \pmod{16}$ so $13^4 \equiv (-3)^4 \equiv 81 \equiv 1 \pmod{16}$\\
* $11 \equiv -5 \pmod{16}$ so $11^4 \equiv (-5)^4 \equiv 625 \equiv 1 \pmod{16}$\\

Again, the difference is divisible by 16:\\

* $13^4 - 11^4 \equiv 1 - 1 \equiv 0 \pmod{16}$\\
Finally, let's check divisibility by 32:\\
* $13^2 \equiv 9 \pmod{32}$ so $13^4 \equiv 9^2 \equiv 81 \equiv 17 \pmod{32}$\\
* $11^2 \equiv 5 \pmod{32}$ so $11^4 \equiv 5^2 \equiv 25 \pmod{32}$\\
This time, the difference is not divisible by 32:\\

* $13^4 - 11^4 \equiv 17 - 25 \equiv -8 \pmod{32}$\\
Since the difference is divisible by 16 but not by 32, the largest power of 2 that divides $13^4 - 11^4$ is $\boxed{\textbf{(C)}\ 32}$. \\
\end{minipage}
}

\colorbox{green!20}{
\begin{minipage}{\textwidth}

\textbf{Please output YES if the new solution leads to the same result as the reference solutions; otherwise, output NO.}
\end{minipage}
}

The solution by Gemini-1.5-Pro is correct, as verified by all \textbf{LLM Evaluators}. After the correctness evaluation, the prompt for novelty evaluation is used.

\clearpage
\subsubsection{Prompt 3. Novelty Evaluation}
\textbf{Task:} The LLM is asked to evaluate whether a newly generated solution is novel compared to existing reference solutions.\\

\colorbox{green!20}{
\begin{minipage}{\textwidth}
\textbf{Criteria for evaluating the novelty of a new mathematical solution include:}
1. If the new solution used to arrive at the solutions is fundamentally different from reference solutions, such as algebraic manipulation versus geometric reasoning, it can be considered novel;\\
2. Even if the final results are the same, if the intermediate steps or processes involved in reaching those solutions vary significantly, the new solution can be considered novel;\\
3. If the new solution relies on different assumptions or conditions, it should be considered novel;\\
4. A solution might generalize to a broader class of problems, while another solution might be specific to certain conditions. In such cases, they are considered distinct;\\
5. If the new solution is significantly simpler or more complex than the others, it can be regarded as essentially novel, even if they lead to the same result.\\

\textbf{Given the following mathematical problem:}\\
What is the largest power of $2$ that is a divisor of $13^4 - 11^4$?\\
$\textbf{(A)}\mbox{ }8\qquad \textbf{(B)}\mbox{ }16\qquad \textbf{(C)}\mbox{ }32\qquad \textbf{(D)}\mbox{ }64\qquad \textbf{(E)}\mbox{ }128$\\

\textbf{Reference solutions:}\\
\textbf{Solution 1:}\\
First, we use difference of squares on $13^4 - 11^4 = (13^2)^2 - (11^2)^2$ to get $13^4 - 11^4 = (13^2 + 11^2)(13^2 - 11^2)$. Using difference of squares again and simplifying, we get $(169 + 121)(13+11)(13-11) = 290 \cdot 24 \cdot 2 = (2\cdot 8 \cdot 2) \cdot (3 \cdot 145)$. Realizing that we don't need the right-hand side because it doesn't contain any factor of 2, we see that the greatest power of $2$ that is a divisor $13^4 - 11^4$ is $\boxed{\textbf{(C)}\ 32}$.\\

\textbf{Solution 2:}\\
Just like in the above solution, we use the difference-of-squares factorization, but only once to get $13^4-11^4=(13^2-11^2)(13^2+11^2).$ We can then compute that this is equal to $48\cdot290.$ Note that $290=2\cdot145$ (we don't need to factorize any further as $145$ is already odd) thus the largest power of $2$ that divides $290$ is only $2^1=2,$ while $48=2^4\cdot3,$ so the largest power of $2$ that divides $48$ is $2^4=16.$ Hence, the largest power of $2$ that is a divisor of $13^4-11^4$ is $2\cdot16=\boxed{\textbf{(C)}~32}.$\\

\textbf{Solution 3:}\\
Let $n=13^4-11^4.$ We wish to find the largest power of $2$ that divides $n$.\\
Denote $v_p(k)$ as the largest exponent of $p$ in the prime factorization of $n$. In this problem, we have $p=2$. \\
By the Lifting the Exponent Lemma on $n$, 
\[v_2(13^4-11^4)=v_2(13-11)+v_2(4)+v_2(13+11)-1\]
\[=v_2(2)+v_2(4)+v_2(24)-1\]
\[=1+2+3-1=5.\]

Therefore, exponent of the largest power of $2$ that divids $13^4-11^4$ is $5,$ so the largest power of $2$ that divides this number is $2^5=\boxed{\textbf{(C)} 32}$.\\

\textbf{Solution 4:}\\
We can simply take 13 to the 4th power, which is 28561. We subtract that by 11 to the 4th power, which is 14641 (You can use Pascal's Triangle to find this). Finally, subtract the numbers to get 13920. \\
To test the options, since we need the largest one, we can go from top down. Testing, we see that both D and E are decimals,
and 32 works. So, our answer is $\boxed{\textbf{(C)}~32}.$\\

\textbf{New solution:}\\
We can solve this problem by considering the remainders when powers of 2 divide 13 and 11. \\
\end{minipage}
}

\colorbox{green!20}{
\begin{minipage}{\textwidth}
* Notice that $13 \equiv 1 \pmod{8}$ (13 leaves a remainder of 1 when divided by 8). \\
* Therefore, $13^4 \equiv 1^4 \equiv 1 \pmod{8}$.\\
* Similarly, $11 \equiv 3 \pmod{8}$, so $11^4 \equiv 3^4 \equiv 81 \equiv 1 \pmod{8}$.\\
Since both $13^4$ and $11^4$ have the same remainder (1) when divided by 8, their difference is divisible by 8:\\

*  $13^4 - 11^4 \equiv 1 - 1 \equiv 0 \pmod{8}$\\

Now, let's check divisibility by 16:\\

* $13 \equiv -3 \pmod{16}$ so $13^4 \equiv (-3)^4 \equiv 81 \equiv 1 \pmod{16}$\\
* $11 \equiv -5 \pmod{16}$ so $11^4 \equiv (-5)^4 \equiv 625 \equiv 1 \pmod{16}$\\

Again, the difference is divisible by 16:\\

* $13^4 - 11^4 \equiv 1 - 1 \equiv 0 \pmod{16}$\\

Finally, let's check divisibility by 32:\\

* $13^2 \equiv 9 \pmod{32}$ so $13^4 \equiv 9^2 \equiv 81 \equiv 17 \pmod{32}$\\
* $11^2 \equiv 5 \pmod{32}$ so $11^4 \equiv 5^2 \equiv 25 \pmod{32}$\\

This time, the difference is not divisible by 32:\\

* $13^4 - 11^4 \equiv 17 - 25 \equiv -8 \pmod{32}$\\

Since the difference is divisible by 16 but not by 32, the largest power of 2 that divides $13^4 - 11^4$ is $\boxed{\textbf{(C)}\ 32}$. \\

\textbf{Please output YES if the new solution is a novel solution; otherwise, output NO.}
\end{minipage}
}

The new solution by Gemini-1.5-Pro is distinct from all reference solutions, which has been classified as a novel solution by all three \textbf{LLM Evaluators}. The reasoning and decision why Claude-3.5-Connect classified this solution as a novel solution are illustrated as follows.\\

\colorbox{red!20}{
\begin{minipage}{\textwidth}
To evaluate the novelty of the new solution, let's compare it against the criteria provided and the reference solutions:\\

1. The new solution uses modular arithmetic and congruences, which is fundamentally different from the algebraic manipulations and factorizations used in the reference solutions. This approach is distinct from the methods used in the reference solutions.\\

2. The intermediate steps in the new solution involve checking remainders when dividing by powers of 2, which is significantly different from the processes in the reference solutions.\\

3. The new solution relies on the properties of modular arithmetic and congruences, which are different assumptions from those used in the reference solutions.\\

4. While the new solution doesn't necessarily generalize to a broader class of problems, it does provide a different perspective on approaching such problems.\\

5. The new solution is neither significantly simpler nor more complex than the reference solutions, but it does offer a distinct approach to solving the problem.\\
Given these considerations, particularly the fundamentally different approach (criterion 1) and the distinct intermediate steps (criterion 2), we can conclude that this new solution meets the criteria for novelty.\\

\textbf{Output: YES}
\end{minipage}
}

\end{document}